\documentclass[nohyperref]{article}

\usepackage{amsmath,amsfonts,bm}

\def\eqref#1{equation~\ref{#1}}

\def\1{\bm{1}}

\DeclareMathAlphabet{\mathsfit}{\encodingdefault}{\sfdefault}{m}{sl}
\SetMathAlphabet{\mathsfit}{bold}{\encodingdefault}{\sfdefault}{bx}{n}

\usepackage{microtype}
\usepackage{graphicx}
\usepackage{subfigure}
\usepackage{booktabs} %

\usepackage{hyperref}

\usepackage[accepted]{icml2023}

\usepackage{amsmath}
\usepackage{amssymb}
\usepackage{mathtools}
\usepackage{amsthm}

\usepackage[capitalize,noabbrev]{cleveref}

\theoremstyle{plain}

\theoremstyle{definition}

\theoremstyle{remark}

\usepackage{nicematrix}
\usepackage{enumitem}

\usepackage{xspace}
\usepackage{lipsum}
\newcommand{\p}[1]{\vspace{1mm}\noindent\textbf{#1}}

\newcommand{\ie}{\textit{i}.\textit{e}.}
\newcommand{\eg}{\textit{e}.\textit{g}.}

\usepackage{multirow}
\usepackage{bbm}

\icmltitlerunning{Emergent Agentic Transformer from Chain of Hindsight Experience}
\newcommand{\ours}{{Agentic Transformer}\xspace}
\newcommand{\oursabb}{{AT}\xspace}
\newcommand{\oursdata}{{chain of hindsight experience}\xspace}

\begin{document}

\twocolumn[
\icmltitle{
Emergent Agentic Transformer from Chain of Hindsight Experience
}

\begin{icmlauthorlist}
\icmlauthor{Hao Liu}{}
\icmlauthor{Pieter Abbeel}{} \\
\vspace{0.1em}
University of California, Berkeley 
\end{icmlauthorlist}

\icmlcorrespondingauthor{Hao Liu}{hao.liu@cs.berkeley.edu}

\icmlkeywords{Machine Learning, ICML}

\vskip 0.3in
]

\printAffiliationsAndNotice{}  %

\begin{abstract}
Large transformer models powered by diverse data and model scale have dominated natural language modeling and computer vision and pushed the frontier of multiple AI areas. In reinforcement learning (RL), despite many efforts into transformer-based policies, a key limitation, however, is that current transformer-based policies cannot learn by directly combining information from multiple sub-optimal trials. In this work, we address this issue using recently proposed chain of hindsight to relabel experience, where we train a transformer on a sequence of trajectory experience ascending sorted according to their total rewards. 
Our method consists of relabelling target return of each trajectory to the maximum total reward among in sequence of trajectories and training an autoregressive model to predict actions conditioning on past states, actions, rewards, target returns, and task completion tokens, the resulting model, Agentic Transformer (AT), can learn to improve upon itself both at training and test time. As we show on D4RL and ExoRL benchmarks, to the best our knowledge, this is the first time that a simple transformer-based model performs competitively with both temporal-difference and imitation-learning-based approaches, even from sub-optimal data. Our Agentic Transformer also shows a promising scaling trend that bigger models consistently improve results.
\end{abstract}

\section{Introduction}
Large transformer~\citep{ashish2017attention} models have substantially advanced the state-of-the-art across a variety of domains, including natural language processing tasks~\citep{devlin2018bert, brown2020language, liu2019roberta}, computer vision~\citep{dosovitskiy2020image, alayrac2022flamingo}, and code generation~\citep{lewkowycz2022solving, chen2021evaluating}.

\begin{figure}[!t]
    \centering
    \includegraphics[width=0.48\textwidth]{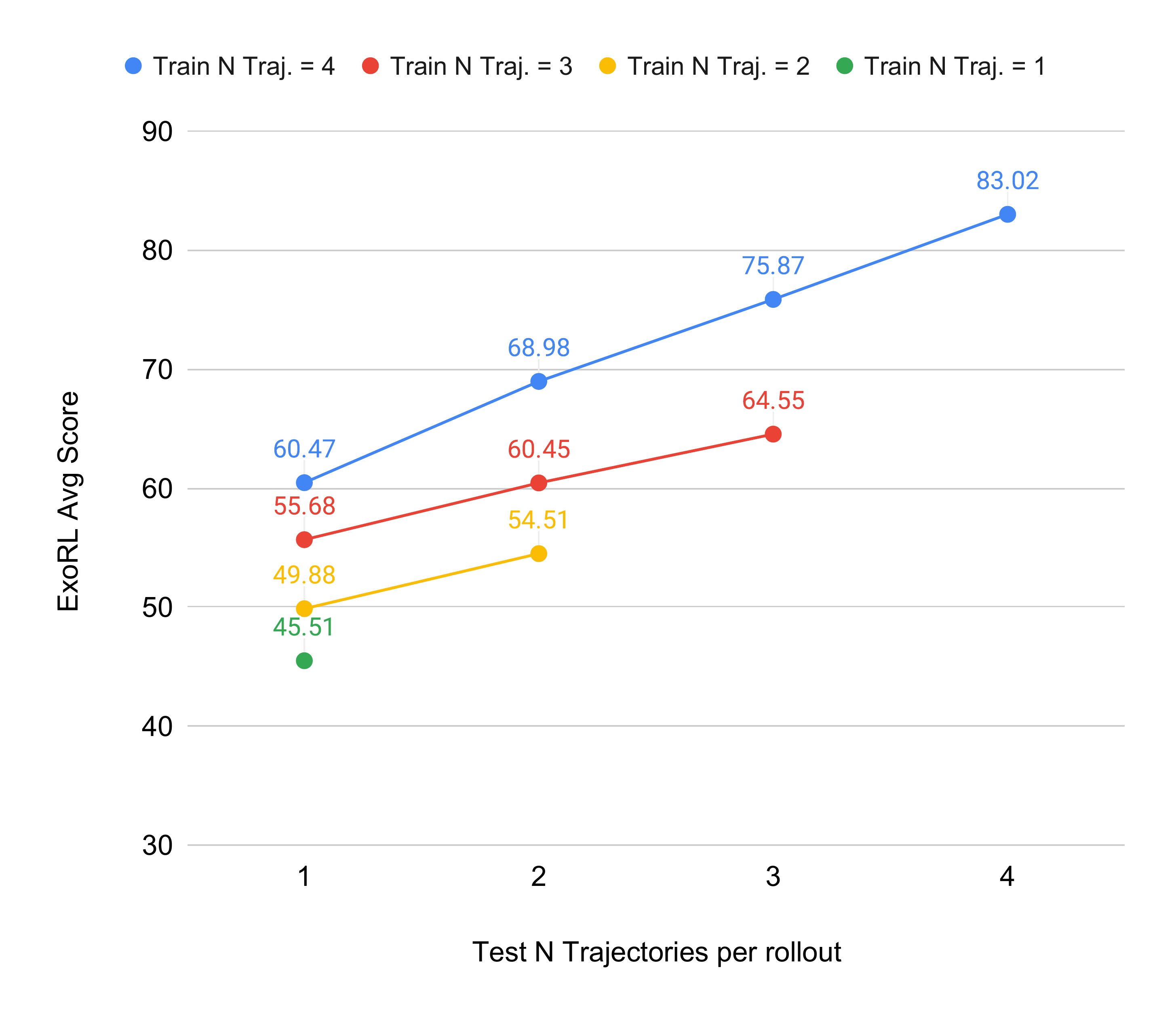}
    \vspace{-2.0em}
    \caption{\ours can automatically improve its performance at evaluation time by rollouting more trajectories in a trial-and-error manner. The scaling improves with both more chain of hindsight training sequences.}
    \vspace{-0.5em}
    \label{fig:agency}
\end{figure}

Despite the successes, a key limitation is that these models are not agentic, \ie\, they cannot interact with the real world to accomplish tasks like a robot.
Reinforcement learning (RL), on the other hand, in principle is designed for building interactive agents. However, conventional RL algorithms are limited to small models (\eg, an MLP with two layers) and are difficult to train and scale~\citep[see e.g.][]{andrychowicz2020matters}. The difficulty of scaling the model size in conventional RL algorithms make it difficult to take advantage of large Transformer models.

In order to combine Transformer with decision-making, there have been lots of efforts in attempting to cast RL from offline data as a sequence modeling problem~\citep{chen2021decision, laskin2022context, reed2022generalist}. For instance, DT~\citep{chen2021decision} proposes to train a Transformer to autoregressively predict action sequences based on sequences of returns-to-go and states. 

Despite the progress made, existing Transformer based decision-making models cannot learn by directly combining information from multiple sub-optimal trials, in fact, they require high-return data to achieve high return~\citep[see e.g.][]{chen2021decision, laskin2022context, yarats2022don}, indicating the lack of extrapolation ability besides the imitation learning ability. 
This limits the wider applicability of transformer-based policies since high return data are not easily available in most important real-world domains, \eg, health care and industry robots.

To resolve these issues, we first hypothesize that the fact that existing Transformer based decision-making models under-perform TD-learning approaches and lack of extrapolation is due to the fact that during training and inference, the model can only do one trial.
Our key observation is that one ability humans have, unlike the current generation of models, is to learn almost as much from achieving an undesired outcome as from the desired one. 
We take the approach \emph{chain of hindsight} introduced in~\citet{liu2023chain} which proposes to condition language model on positive indicator and negative example to predict positive example, and vice versa. 
The idea applies to learn decision making -- imagine learning basketball and attempting a shot that misses the net on the right. Existing models conclude that the sequence of performed actions don't result in success, and little is learned.
It is however possible to chain another attempt's sequence of actions which missed even more far away with this sequence of actions, as if this sequence of actions would be a successful second attempt if the goal is placing the ball closer to the net.

In this paper, we propose to train Transformer to perform exactly this kind of reasoning. 
Through training on \emph{chain of hindsight} experience, the resulting model is named as \ours (\oursabb).
Not only does \ours improve the performance on learning from high return data, but more importantly, it makes learning possible even if the data is far from being optimal.
Our approach is based on training a decoder-only Transformer~\citep{radford2018improving, radford2019language, brown2020language} which takes as input not only the current episode, but also multiple episodes whose returns are lower than current episode's return and are ascending sorted according to their returns.  
The pivotal idea behind \ours is to replay each episode with a variable number (\eg, randomly choose between 0 and 4) of episodes to form \oursdata, as if the model was trying to improve from previous episode(s) to current episode.

\ours achieves state-of-the-arts on standard RL benchmarks including D4RL~\citep{fu2020d4rl} and ExoRL~\citep{laskin2021urlb, yarats2022don}. 
\ours can learn by directly combining information from multiple sub-optimal trials and being able to improve itself through multiple trials at test time.
Our experiments show that AT scales well in both model size and the length of \oursdata, indicating further improvement could be possible by scaling up model and data.

\begin{figure*}[!t]
    \centering
    \includegraphics[width=0.99\textwidth,keepaspectratio]{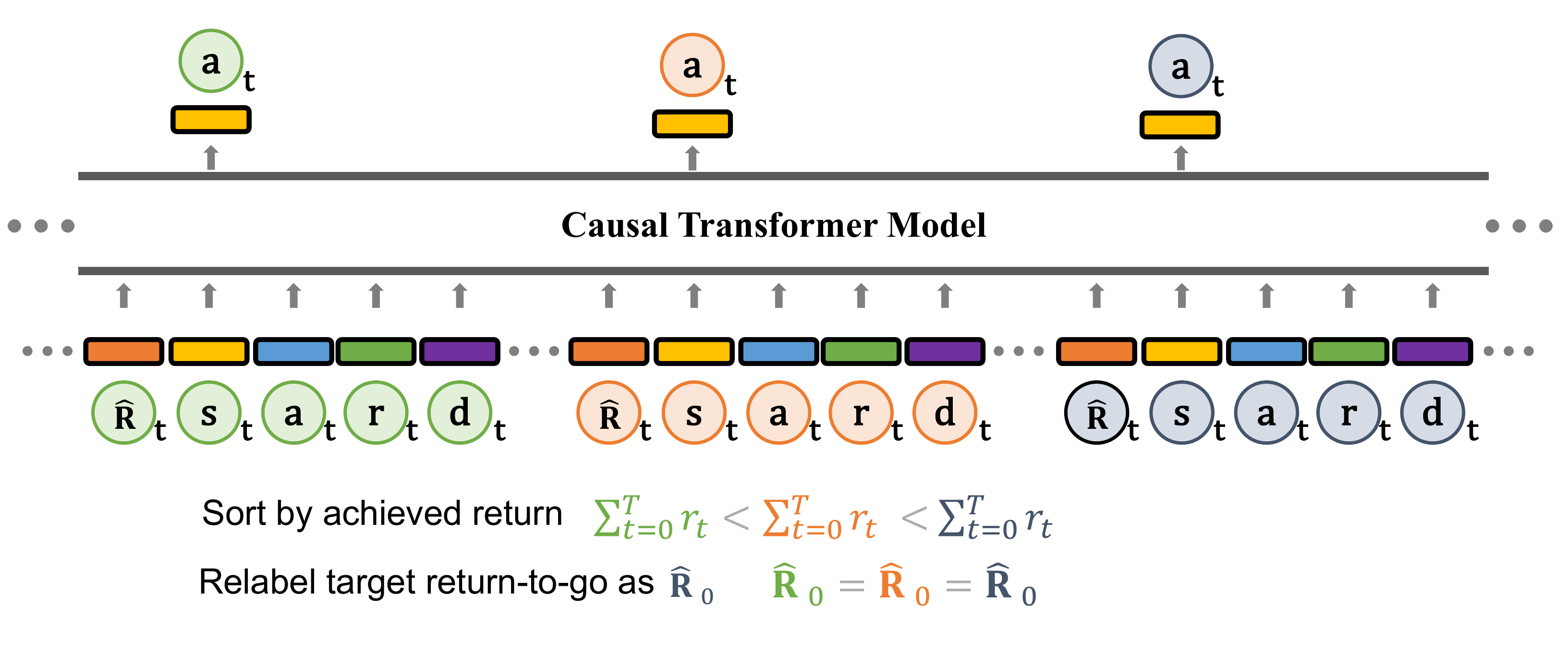}
    \vspace{-1em}
    \caption{\textbf{Agentic Transformer}. The input sequence consists of multiple episodes ascending sorted according to their total rewards.
    The \underline{initial desired return} $\widehat{R}_0$ of all trajectories are set to the maximum total rewards among all trajectories. 
    For each trajectory, the return-to-go is updated using rewards in \underline{the same trajectory}: $\widehat{R}_t = \widehat{R}_0 - \sum_{j=0}^t r_j$.
    The \underline{task completion token} $d$ indicates whether achieved cumulative rewards in a trajectory is larger than desired target return( Equation~\ref{eq:task_completion}), this gives model feedback on past trajectories and help steer model to try to reach target return in next trajectory at test time.
    States, actions, rewards, returns-to-go, and task completion are fed into modality speciﬁc linear embeddings and a positional episodic timestep encoding is added. Tokens are fed into a GPT architecture which predicts actions autoregressively using a causal self-attention mask.
    At \texttt{training time:}
    The model is trained to predict action tokens in the \underline{last (best) trajectory} conditioning on past trajectories, states, actions, returns-to-go and task completion tokens.
    At \texttt{testing time:}
    The model predicts action autoregressively across multiple trajectories.
    }
    \label{fig:model}
\end{figure*}

\section{Preliminaries}

\subsection{Reinforcement Learning}
We consider learning problem in the context of a Markov Decision Process (MDP) represented by the tuple ($\mathcal{S}$, $\mathcal{A}$, $P$, $\mathcal{R}$). The MDP tuple consists of states $s \in \mathcal{S}$, actions $a \in \mathcal{A}$, transition dynamics $P(s'|s,a)$, and a reward function $r = \mathcal{R}(s,a)$.
To describe the state, action, and reward at time step $t$, the notations $s_t$, $a_t$, and $r_t=\mathcal{R}(s_t, a_t)$ are used. A trajectory is a sequence of states, actions, and rewards and is denoted by $\tau = (s_0, a_0, r_0, s_1, a_1, r_1, \ldots, s_T, a_T, r_T)$. The return of a trajectory at time step $t$, $R_t = \sum_{t'=t}^T r_{t'}$, is calculated as the sum of future rewards from that time step.
The goal of reinforcement learning is to find a policy that maximizes the expected return $\mathbb{E}\Bigl[\sum_{t=1}^T r_t\Bigr]$ in an MDP.
In supervised or offline reinforcement learning, data is obtained from a fixed limited dataset of trajectory rollouts from arbitrary policies, instead of from environment interactions. This setting eliminates the ability of the agents to explore the environment and gather additional feedback. Conventional datasets either consist mainly of high quality, near optimal trajectories like in D4RL~\citep{fu2020d4rl} which are obtained by running trained expert policies or by storing the experience of training an expert policy, or mainly consist of diverse, exploratory and sub-optimal trajectories like in ExoRL~\citep{yarats2022don} where trajectories are collected through unsupervised exploration algorithms.

\subsection{Transformers}
The Transformer~\citep{ashish2017attention} architecture consists of multiple layers of self-attention operation and MLP.
The self-attention begins by projecting input data $X$ with three separate matrices onto $D$-dimensional vectors called queries $Q$, keys $K$, and values $V$. These vectors are then passed through the attention function:
\begin{align}
    \text{Attention}(Q,K,V) = \text{softmax} (Q K^T / \sqrt{D}) V.
\end{align}
The $QK^T$ term computes an inner product between two projections of the input data $X$. The inner product is then normalized and projected back to a $D$-dimensional vector with the scaling term $V$. Transformers~\citep{ashish2017attention, devlin2018bert, brown2020language} utilize self-attention as a core part of the architecture to process sequential data such as text sequences. Transformers are usually pre-trained with a self-supervised objective. Common prediction tasks include predicting randomly masked out tokens~\citep{devlin2018bert} or applying a causal mask and predicting the next token~\citep{radford2018improving, brown2020language}. 
The GPT architecture \citep{radford2018improving} replaces the summation/softmax over the $n$ tokens with only the previous tokens in the sequence ($j \in [1, i]$), enabling autoregressive generation by using causal self-attention mask.
In this work, we use the GPT architecture because we need to do autoregressive generation at test time.

\subsection{Transformer based Behavior Cloning}
We refer to the family of methods that treat Reinforcement Learning from offline data as a sequential prediction problem as Transformer based behavior cloning.
Rather than learning a value function from offline data, this family of works focus on extracting policies by predicting actions in the offline data (\ie\, behavior cloning) with an autoregressive sequence model and either return conditioning~\citep{chen2021dt, laskin2022context, lee2022multi} or filtering out suboptimal data~\citep{reed2022generalist} or training masked sequence model bypredicting masked states and actions tokens~\citep{liu2022masked, carroll2022unimask}.

\section{Method}

In this section, we present Agentic Transformer (AT), which models \oursdata trajectories autoregressively based on Transformer archiecture, as summarized in Figure \ref{fig:model} and Algorithm \ref{alg:main}.

\textbf{Chain of hindsight Experience.}
The key factors that influenced our decision on how to represent trajectories are: (1) the ability of transformers to uncover meaningful patterns from multiple trajectories sampled from arbitrary offline data, and (2) the capacity to produce actions conditionally during evaluation and improve itself conditions on collected experience.
Modeling rewards is a nontrivial task, therefore, we aimed to have the model generate actions based on the \emph{future} desired returns, similar to previous works~\citep[\eg,][]{chen2021decision, laskin2022context}, rather than relying on past rewards.
We feed the model with the initial target returns-to-go $\widehat{R}_0$ and update $\widehat{R}_t = \widehat{R}_0 - \sum_{j=0}^t r_j$ using rewards.
We also feed the model with a completion token $d$ that indicates whether the achieved cumulative rewards in a trajectory are larger than or equal to desired returns-to-go, specifically
\begin{align}
d_T  = \mathbbm{1}\left(\sum_{j=0}^T r_j \geq \widehat{R}_0\right) \quad d_i = 0, ~\forall i \in [1, T-1],
\label{eq:task_completion}
\end{align}
where $\mathbbm{1}$ is indicator function.
This leads to the following trajectory representation which is amenable to autoregressive training and generation:
\begin{align}
    \tau & = \left(\widehat{R}_0, s_0, a_0, r_0, d_0, \dots, \widehat{R}_T, s_T, a_T, r_T, d_T\right) \nonumber  \\
    &\quad \text{where} \quad \widehat{R}_t = \widehat{R}_0 - \sum_{j=0}^t r_j.
    \label{eq:episode}
\end{align}

Since we want the model to learn to 'stitch' sub-optimal data rather than just imitating optimal data, and at test time we want the model to achieve desired target return through multiple trajectories of trial-and-errors, 
we construct a chain of hindsight experience for the model to learn to improve even from sub-optimal data and learning to self-improve during test time. To achieve this, we take the approach called \emph{chain of hindsight}~\citep{liu2023chain} which trains language model from human feedback by conditioning on positive indicator and negative rated example to predict corresponding positive rated example. 
And adapt it to decision making by replaying each episode with a variable number (\eg, randomly choose between 0 and 4) of episodes to form \emph{chain of hindsight} experience, as if the model was trying to improve from previous episode(s) to current episode.

This leads to the following chain of hindsight trajectory representation:
\begin{align}
    & s = \left(\tau^1, \tau^2 \dots, \tau^n \right) \\
    & \text{where} \nonumber \\
    & \,\,\,\, \tau^i = \left(\widehat{R}^i_0, s^i_0, a^i_0, r^i_0, d^i_0, \dots, \widehat{R}^i_T, s^i_T, a^i_T, r^i_T, d^i_T\right) \\
    & s.t. \nonumber \\
    & \,\,\,\, \sum_{t=1}^T r^0_t \leq \sum_{t=1}^T r^1_t \leq \dots \leq \sum_{t=1}^T r^n_t \label{eq:chain_order} \\
    & \,\,\,\, \widehat{R}^i_0 = \sum_{t=1}^T r^n_t \quad \forall~ 1 \leq i \leq n \label{eq:chain_target} \\
    & \,\,\,\, \widehat{R}^i_t = \widehat{R}^i_0 - \sum_{j=0}^t r^i_j \quad \forall~ 1 \leq i \leq n,
    \label{eq:chain_rtg}
\end{align}
Equation~\ref{eq:chain_order} states the ordering requirement, meaning that trajectories are ascending sorted according to their total reward.
Equation~\ref{eq:chain_target} sets the \emph{hindsight} target: for all $n$ trajectories, initial target equals to trajectory $n$'s total reward.
Equation~\ref{eq:chain_rtg} updates  returns-to-go using trajectory reward. 

At test time, we can specify the desired performance (e.g. 1 for success or 0 for failure), as well as the environment starting state, and the conditioning information to initiate generation.
After executing the generated action for the current state, we decrement the target return by the achieved reward and repeat until episode termination. 
If the target return is not achieved, the model starts a new episode and continues interacting with the environment until the maximum episode number is reached.

\textbf{Architecture.}
We feed the $n$ trajectories into \ours, this results in a total of $5\times n \times T$ tokens, with one token for each of the five modalities: returns-to-go, state, action, reward, and completion. To create the token embeddings, a linear layer is trained for each modality which transforms the raw inputs into the desired embedding dimension, followed by layer normalization~\citep{ba2016layer}.
In addition to this, an embedding for each time step is also learned and added to the tokens, which is distinct from the standard positional embedding used in transformers where one time step is represented by five tokens. Finally, the tokens are processed by a GPT model~\citep{radford2018improving} that predicts future action tokens through autoregressive modeling.

\textbf{Training and Test.}
We are given a dataset of offline trajectories.
We sample minibatches of trajectories from the dataset.
The model predicts the action token $a_t$ given the input token $s_t$, and the prediction is evaluated with either cross-entropy loss or mean-squared error, depending on whether the actions are discrete or continuous. The losses from each time step are averaged.
Note that only the action tokens $a_t$ from the last trajectory $\tau^n$ are used for loss calculation.
While it's feasible to predict other tokens or use other trajectories in the training process, we didn't observe improvements in performance and consider it as a potential area for future research.
At test time, following standard practice in NLP, we cache key and query during autoregressive decoding to speed up inference.
For transformer based models DT and AT, at test time we rollout the model with $n$ trajectories, irregardless cases when $d_T=1$ \ie\,  desired target return is achieved, and report the largest return among $n$ trajectories. 
For DT the maximum return is achieved at the 1st trajectory while AT improves itself along the trajectory sequence and achieves higher return with more trajectories. 
The model sizes are shown in Table~\ref{tab:model_size}, base is used by default unless otherwise mentioned.
Since in our default configuration $n=4$, and $T$ is typically 1000 in D4RL and ExoRL, total sequence length is 20,000 which uses a large amount of memory for large models. To address this issue, we implement \ours using data parallelism on batch dimension and model parallelism on sequence dimension. 
By doing so, we can easily scale \ours across multiple GPUs or TPUs. 
The code of \ours will be made publicly available for future research.
\begin{algorithm}[!htbp]
\caption{Training \ours}
\label{alg:main}
\begin{algorithmic}
\small
   \STATE {\bfseries Required:} Dataset of Trajectories, Transformer Model
   \STATE {\bfseries Required:} Max Iterations $m$, Max Number of trajectories in chain of hindsight experience $n$
   \STATE Initialize
   \FOR{$i=1$ {\bfseries to} $m-1$}
   \STATE Randomly sample $j$ from 1 to $n$
   \STATE Randomly sample $j$ episodes from dataset
   \STATE Compute returns-to-go $\hat{R}$ for all steps for each episode 
   \STATE Sort $j$ episodes ascending according to their returns
   \STATE Let $\hat{R}_\text{max}$ be the return of the last episode
   \STATE For each other episode, recomputing its returns-to-go by setting $\hat{R}_0 = \hat{R}_\text{max}$ 
   \STATE Concatenate $j$ episodes as a sequence
   \STATE Train Transformer to predict next action token (see Figure~\ref{fig:model}).
   \ENDFOR
\end{algorithmic}
\end{algorithm}

\begin{table}[!htbp]
\begin{center}
\small
\begin{NiceTabular}{ lcccc } 
\toprule
Model & Layers   & \# of heads   &  $d_\textrm{model}$ & Batch size  \\ 
\midrule
Small & $2$ & $4$  & $64$ & $256$ \\
Base & $4$ & $8$  & $256$ & $256$ \\
Large & $6$ & $16$  & $512$ & $256$ \\
XLarge & $8$ & $16$  & $512$ & $256$ \\
\bottomrule
\end{NiceTabular}
\end{center}
\vspace{-0.6em}
\caption{Architecture details of different sized models used in \ours. 
We list the number of layers, $d_\textrm{model}$, the number of attention heads and attention head size, training batch size, and sequence length.
The feed-forward size $d_\textrm{ff}$ is always $4 \times d_\textrm{model}$ and attention head size is always 16.
}
\label{tab:model_size}
\end{table}

\begin{figure*}[t]
    \centering
    \setlength{\tabcolsep}{0pt}
    \begin{NiceTabular}[t]{cc}
    \includegraphics[width=.49\textwidth]{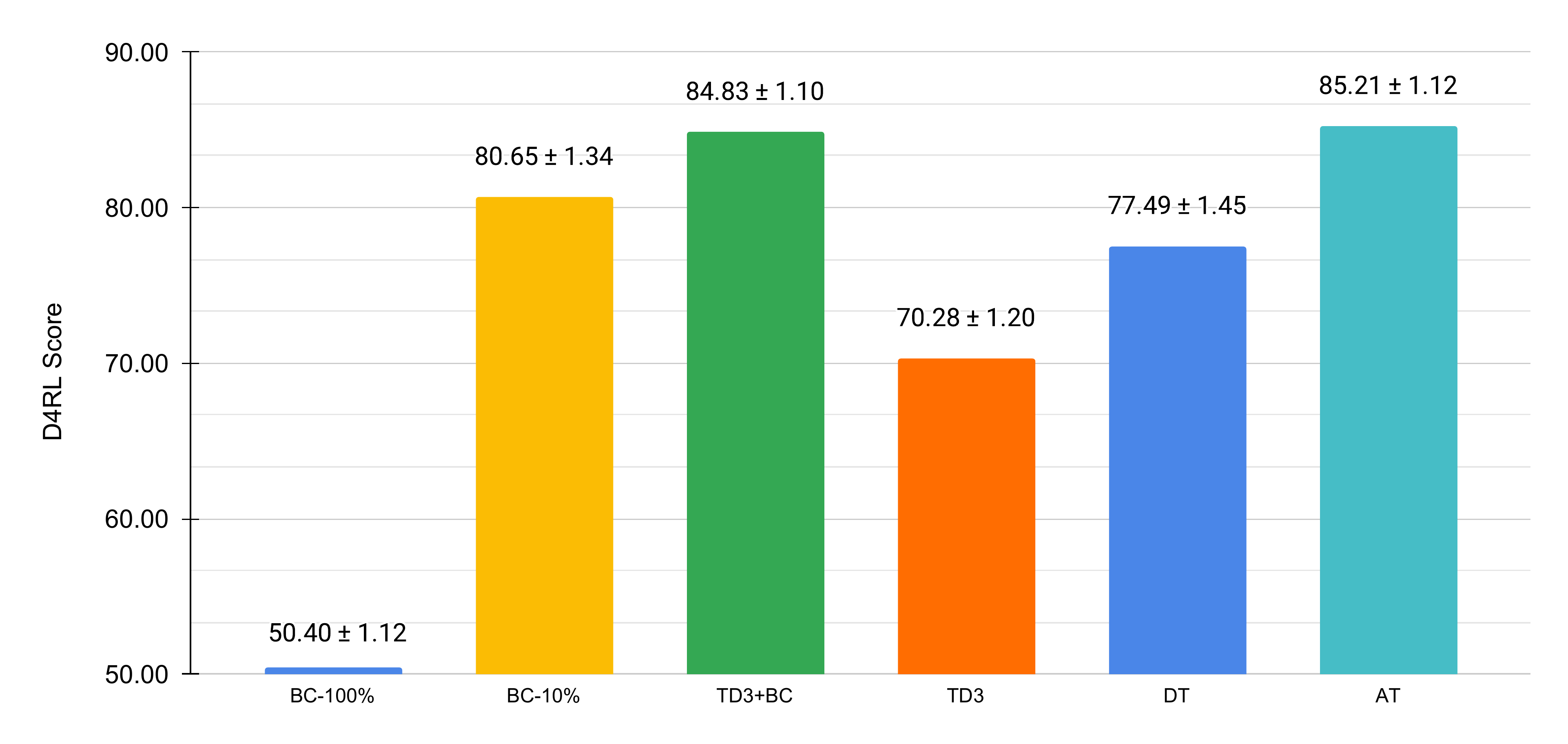}  & \includegraphics[width=.49\textwidth]{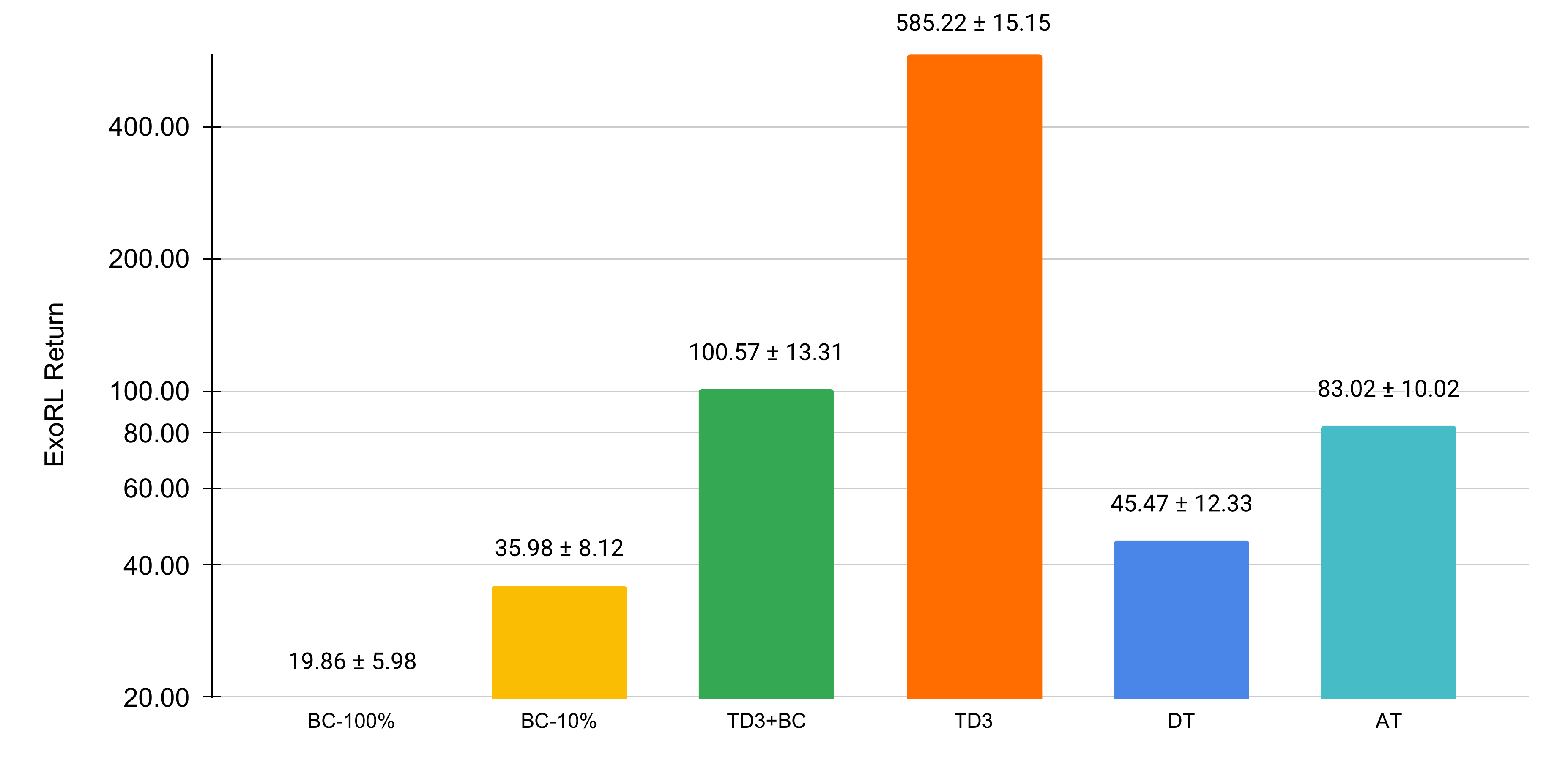} \\
    \end{NiceTabular}
    \vspace{-0.5em}
    \caption{\ours performs competitively with both temporal-difference based and imitation-learning based approaches in ExoRL as well as D4RL tasks. \textbf{Left}. Tasks average performance on D4RL.  
    \textbf{Right}. Tasks average performance on ExoRL. 
    We report the mean and variance for three seeds. 
    }
    \label{fig:bench_teaser}
\end{figure*}

\begin{table*}[t]
\centering
\caption{
Results for D4RL datasets.
We report the mean and variance for three seeds.
Using \oursdata, our Agentic Transformer (AT) outperforms both supervised learning (BC) and Transformer (DT) and performs competitively with conventional RL algorithms (TD3+BC, TD3) on almost all tasks}
\vspace{0.5em}
\vspace{0.5em}
\begin{NiceTabular}{lr||rrrrc}\toprule
\textbf{Dataset} &\textbf{Environment} &BC-10\% &TD3+BC &TD3 &DT &Agentic Transformer (AT) \\\midrule
Medium-Expert &HalfCheetah &94.11 &\bf 96.59 &87.60 &93.40 &95.81 $\pm$ 0.25 \\
Medium-Expert &Hopper &113.13 &113.22 &98.41 &111.18 &\bf 
 115.92 $\pm$ 1.26 \\
Medium-Expert &Walker &109.90 &112.21 &100.52 &108.71 &\bf 114.87 $\pm$ 0.56 \\
Medium &HalfCheetah &43.90 &\bf 48.93 &34.60 &42.73 &45.12 $\pm$ 0.34 \\
Medium &Hopper &73.84 &\bf 70.44 &56.98 &69.42 &\bf 70.45 $\pm$ 0.45 \\
Medium &Walker &82.05 &86.91 &70.95 &74.70 &\bf 88.71 $\pm$ 0.55 \\
Medium-Replay &HalfCheetah &42.27 &45.84 &38.81 &40.31 &\bf 46.86 $\pm$ 0.33 \\
Medium-Replay &Hopper &90.57 &\bf 98.12 &78.90 &88.74 &96.85 $\pm$ 0.41 \\
Medium-Replay &Walker &76.09 &91.17 &65.94 &68.22 &\bf 92.32 $\pm$ 1.21 \\ \cmidrule{1-7}
\multicolumn{2}{c}{\textbf{Total Average}} &80.65 &84.83 &70.30 &77.49 &\bf 85.21 \\
\bottomrule
\end{NiceTabular}
\label{tab:d4rl}
\end{table*}

\section{Experiments}

\p{Dataset: D4RL.}
In this section, we consider the continuous control tasks from the D4RL benchmark~\citep{fu2020d4rl}.
The different dataset settings are described below.
\begin{enumerate}
    \item Medium: 1 million timesteps generated by a ``medium'' policy that performs approximately one-third as well as an expert policy.
    \item Medium-Replay: it contains the replay buffer of an agent trained to the performance of a medium policy.
    \item Medium-Expert: each task consists of one million timesteps generated by the medium policy combined with one million timesteps generated by an expert policy.
\end{enumerate}
The dataset are collected from multiple Mujoco environments including HalfCheetah, Hopper, and Walker. 
Since D4RL dataset is collected by conventional RL algorithms, it consists of many high return trajectories that are near expert. 
Therefore, filtered behavior cloning (\eg 10$\%$ BC) often performs similarly or better than specifically designed offline RL algorithms (\eg DT). 
In order to evaluate our method in a more challenging and realistic setting,
we consider ExoRL~\citep{yarats2022don} dataset that only consists of diverse and low return trajectories.

\p{Dataset: ExoRL.} The ExoRL dataset is based on unlabeled exploratory data collected by running unsupervised RL algorithms. 
For each environment, it comes with eight different unsupervised data collection algorithms, taken from from URLB~\citep{laskin2021urlb}. 
The datasets are collected by unsupervised RL and then relabeled using task reward function.
In light of the benefit of scaling up data~\citep{hoffmann2022training}, we opted to use the combination of all datasets for all baselines and our method.
Specifically, for each environment, we combine the datasets collected by eight algorithms~\citep{pathak2017curiosity, pathak2019self, burda2018exploration, liu2021behavior, yarats2021reinforcement, eysenbach2018diversity, lee2019efficient, liu2021aps}. 
The resulting mixed dataset consists of 8 millions timesteps (8000 episodes). 
Since it is collected by unsupervised RL without using task rewards, the dataset is optimized for diversity but is far from optimal task rewards.
The details are referred to the original papers.

\begin{table*}[t]
\centering
\caption{
Results for ExoRL datasets.
We report the mean and variance for three seeds.
Using \oursdata, our Agentic Transformer (AT) outperforms both supervised learning (BC) and Transformer (DT) on almost all tasks, and performs competitively with conventional RL algorithms (TD3+BC, TD3).}
\vspace{1.5em}
\begin{NiceTabular}{lr||rrrrc}\toprule
\textbf{Dataset} &\textbf{Task} &BC-10\% &TD3+BC &TD3 &DT &Agentic Transformer (AT) \\\midrule
All &Walker Stand &52.91 &67.13 &832.10 &34.54 &68.55 \\
All &Walker Run &34.81 &45.83 &387.76 &49.82 &88.56 \\
All &Walker Walk &13.53 &56.73 &897.81 &34.94 &64.56 \\
All &Cheetah Run &34.66 &187.55 &318.41 &67.53 &125.68 \\
All &Jaco Reach &23.95 &167.85 &287.55 &18.64 &52.98 \\
All &Cartpole Swingup &56.82 &78.57 &787.52 &67.56 &97.81 \\ \midrule
\multicolumn{2}{c}{\textbf{Total Average}} &36.11 &100.61 &585.19 &45.51 &83.02 \\
\bottomrule
\end{NiceTabular}
\label{tab:exorl}
\end{table*}

\p{Baselines.}
In this section, we investigate the performance of \ours relative to dedicated offline RL, imitation learning algorithms, and Transformer-based policies. 
In particular, our primary points of comparison are prior Transformer-based policies such as decision transformer since architecture wise \ours is similar them. By comparing with them, we can evaluate the effectiveness of \oursdata and other algorithmic improvements. 
We further compare with model-free offline RL algorithms based on TD-learning, since architecture is fundamentally model-free in nature as well.
Furthermore, TD-learning is the dominant paradigm in RL for sample efficiency and is effective at learning from sub-optimal data. 
By comparing \ours with TD-learning in both high-return and low-return datasets, we can see if our transformer-based policy can do extrapolation. 
We also compare with behavior cloning and variants, since it also involves a likelihood based policy learning formulation similar to ours. 
Our baselines can be categorized as follows:
\begin{itemize}
    \item \textbf{Transformer-based Policy}: these models use transformer to model trajectory sequence and predict action autoregressively. We consider decision transformer (DT)~\citep{chen2021dt} which is shown to be effective on D4RL.
    \item \textbf{TD learning}: most of these methods use an action-space constraint or value pessimism, and will be the most faithful comparison to \ours, representing standard RL methods. We consider state-of-the-art TD3+BC~\citep{fujimoto2021minimalist} which is shown to be effective on D4RL and TD3~\citep{fujimoto2018addressing} which is shown to be effective on ExoRL.
    \item \textbf{Imitation learning and Behavior Cloning}: this regime similarly uses supervised losses for training, rather than Bellman backups. We consider BC-10\%. BC-10\% is shown to be competitive to state-of-the-arts on D4RL. DT also belongs to this category since it is a transformer based return conditioned BC, both are closely related to our model.
\end{itemize}
In total for offline RL we use five algorithms: BC-10\%, TD3+BC, TD3 and DT. We adhere closely to the original hyper-parameter settings for each algorithm, but in several cases we perform hyper-parameter tuning to achieve best possible performance. We train offline RL algorithms for 500k gradient updates and then evaluate by rolling out $10$ episodes in the environment. We report mean and standard error across $3$ random seeds.

\subsection{D4RL results}
On D4RL, scores are normalized so that 100 represents an expert policy, as per~\citet{fu2020d4rl}.
Baselines numbers are reported by the original papers and from the D4RL paper.
\ours surpasses the baselines in a wide range of tasks. Our results are shown in Table \ref{tab:d4rl}.
Overall, \ours achieves strongest results in a majority of the tasks and is competitive with the state of the art in the remaining tasks.

Since TD3+BC and DT are generally the best algorithms in temporal-difference learning and behavior cloning categories, the superior performance of \ours clearly demonstrate the advantages of using \oursdata.

\subsection{ExoRL results}
On ExoRL, we report the cumulative return, as per~\citet{yarats2022don}.
BC, TD3+BC, and TD3 numbers are from the ExoRL paper, DT numbers are run by ourselves.
Our results are shown in Table \ref{tab:exorl}.
\ours achieves the highest scores in a majority of the tasks and is competitive with the state of the art in the remaining tasks.

Since the ExoRL data is significantly more diverse than D4RL because it is collected using unsupervised RL~\citep{laskin2021urlb}, it is found that temporal-difference learning performs best while behavior cloning struggles. 
\ours significantly outperforms behavior cloning approaches BC-10\% and DT, and achieves competitive results with TD learning approaches. 

\begin{figure}[!htbp]
    \centering
    \includegraphics[width=0.48\textwidth,keepaspectratio]{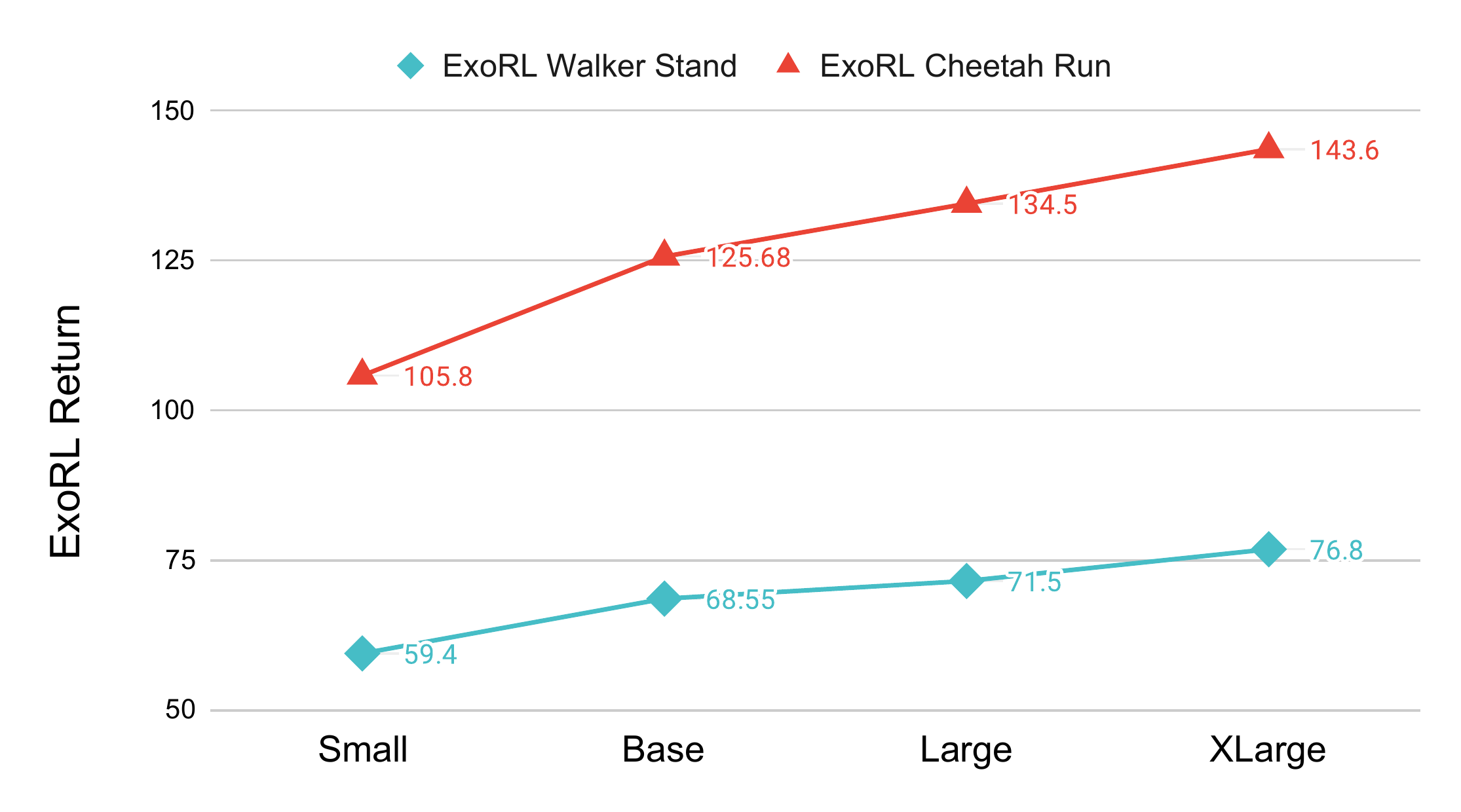}
    \caption{The results of \ours with different model sizes on two ExoRL tasks.}
    \label{fig:model_scaling}
\end{figure}

We further evaluate \ours with different models sizes. 
We select two tasks from ExoRL in order to reduce compute cost incurred by XLarge model size.
Figure~\ref{fig:model_scaling} shows the results. 
\ours improves with larger model size, showing promising scaling behavior.

\subsection{Evaluation of Agency}
At test time, the total rewards of each trajectory in a sequence are reported in Figure~\ref{fig:agency}.
We follow DT's experimental settings and use their target return as initial return-to-go for both DT and AT.

As the number of trajectories increases, the return for AT also increases. In some cases, AT is able to attain the desired target return by the 2nd or 3rd trajectory, resulting in a higher return in the last 4th trajectory. On the other hand, when multiple trajectories are rolled out using DT, the results are poor. DT is unable to produce consistent or higher returns beyond the 1st trajectory.

\subsection{Model Variations}

\begin{table*}[!t]
\centering
\caption{Variations on the \ours and \oursdata. Unlisted values are identical to those of the default configuration. All metrics are averaged over 3 random seeds based on the ExoRL and D4RL benchmarks.
}
\vspace{1.0em}
\scalebox{0.85}{
\begin{NiceTabular}{c||rrrrrrr|cc}\toprule
\textbf{Variants} &\textbf{With 'd'} &\textbf{With 'r'} &\textbf{Hindsight Tgt} &\textbf{Ordered} &\textbf{\# Test Traj} &\textbf{\# Train Traj} &\textbf{All tokens loss} &\textbf{ExoRL Avg} &\textbf{D4RL Avg} \\\midrule
\textbf{Default} &true &true &4th &true &4 &4 &false &83.02 &85.21 \\ \midrule
\multirow{3}{*}{\textbf{(A)}} & & & & & &3 & &76.19 &82.45 \\
& & & & & &2 & &65.47 &80.85 \\
& & & & & &1 & &46.45 &80.26 \\ \midrule
\multirow{4}{*}{\textbf{(B)}} &\multirow{4}{*}{false} & & & &1 & & &57.09 &74.34 \\
& & & & &2 & & &61.92 &73.56 \\
& & & & &3 & & &60.91 &70.88 \\
& & & & &4 & & &61.20 &75.68 \\ \midrule
\textbf{(C)} & &false & & & & & &76.59 &80.43 \\ \midrule
\multirow{3}{*}{\textbf{(D)}} & & &1st & & & & &14.18 &52.33 \\
& & &2nd & & & & &32.29 &65.55 \\
& & &3rd & & & & &58.48 &78.81 \\ \midrule
\textbf{(E)} & & & &false & & & &17.25 &29.78 \\ \midrule
\multirow{3}{*}{\textbf{(F)}} & & & & &1 & & &58.35 &81.48 \\
& & & & &2 & & &74.17 &82.56 \\
& & & & &3 & & &76.29 &84.88 \\ \midrule
\multirow{4}{*}{\textbf{(G)}} & & & & &1 & &true &35.88 &66.45 \\
& & & & &2 & &true &66.30 &71.55 \\
& & & & &3 & &true &73.16 &76.80 \\
& & & & &4 & &true &73.88 &78.88 \\
\bottomrule
\end{NiceTabular}
}
\label{tab:variations}
\vspace{0.5em}
\end{table*}

To evaluate the importance of different components of \ours, we varied our default model in different ways, measuring the change in performance on ExoRL and D4RL benchmarks.  
We present these results in Table~\ref{tab:variations}.  

In Table~\ref{tab:variations} rows (A), we vary the number of training trajectories $n$, keeping the number of testing trajectories constant. Performance improves with the increasing of number of training trajectories, this indicates a promising scaling direction for further improvement. 

In Table~\ref{tab:variations} rows (B), we remove the task completion token 'd' from the input sequence, so the model is trained to 'blindly' learn from hindsight experience. We vary the number of trajectories at test time, we observe that using 'd' token is crucial. While without it \ours still outperforms baselines, the performance degrades significantly compared with default configuration. In addition, without this completion token, the model does not improve with more trajectories at test time, indicating that completion token is important for the model to learn from hindsight experience.

In Table~\ref{tab:variations} rows (C), we observe that removing reward token 'r' has minimal negative effect. This is probably because the model can infer reward token by a simple subtraction from two consecutive returns-to-go tokens. 

In Table~\ref{tab:variations} rows (D), we vary the desired return $\hat{R}_0$. Since default configuration uses 4 trajectories, the default target equals to total reward of last trajectory $\hat{R}_0 = \sum_{t=1}^T r_t^4$. We vary $\hat{R}_0$ to be the total reward of other trajectories. 
We observe that changing this target decreases performance significantly, with the largest decrease happens when $\hat{R}_0$ equals the total reward of the first trajectory. 

In Table~\ref{tab:variations} rows (E), instead of having ordered trajectories $s = \left(\tau^1, \tau^2 \dots, \tau^n \right)$, we randomly shuffle all $\tau$ for each training batch.
We observed significantly worse results, in particularly on ExoRL, this change decreases the performance to only slightly better than BC and DT.

In Table~\ref{tab:variations} rows (F), we evaluate different number of trajectories at test time, we observed a steady better result from using more trajectories at test time. We further observe that although results are better with more trajectories, even using one trajectory, \ours still outperforms Transformer-based policies on both ExoRL and D4RL benchmarks. This suggests that \ours not only learns more than just imitation learning, but also learns to improve upon its own experience.

In Table~\ref{tab:variations} rows (G), we consider applying loss on all trajectories rather than just last trajectory. We observe that it is detrimental to performance, and particularly reduces performance for when the number of test trajectories is small. This suggests that it is best to optimize model towards 'better' behaviors rather than imitating all behaviors.

\section{Related Work}

\subsection{Transformer for Decision-Making}
Prior works explored using Transformers in the context of supervised or offline RL. 
Among them, decision transformer (DT)~\citep{chen2021decision} proposes to model trajectories as sequences and autoregressively predicts action conditioning on desired returns-to-go and past states and actions.
Our model takes input as multiple trajectories and conditions on hindsight information for learning to improve. 
~\citet{chen2021decision} found that DT does not benefit from longer context window and the results saturates at very short context length (\eg, 3-5), possibly due to Markovian environments. Our \ours (\oursabb) models non-Markovian multiple episodes, it shows improved results with longer context length and benefits from Transformers architecture. 
Algorithm distillation (AD)~\citep{laskin2022context} also conditions the model on multiple trajectories, the difference is that AD requires the data to be the experience over the life time of a RL algorithm, while our model can learn from data from any sources. 
Another key difference is our model conditions on hindsight information including hindsight desired returns-to-go and hindsight task completion tokens. 
We observe these algorithm modifications are crucial for superior performance.
Transformer has been explored in learning general world model~\citep{liu2022masked, carroll2022unimask, wu2023masked}, learning from multiple games~\citep{reed2022generalist, lee2022multi}, offline model-based learning~\citep{janner2021offline, liu2022masked}, meta learning~\citep{melo2022transformers, team2023human},
vision-language navigation~\citep{chen2021history, shah2022lm}, robot learning and behavior cloning from noisy demonstrations~\citep{shafiullah2022behavior, cui2022play}, learning from multiple cameras~\citep{seo2022masked}, and language-conditioned imitation learning~\citep{guhur2022instruction, liu2022instruction, shridhar2022cliport, zheng2022vlmbench}.
Since our model is a general decision-making model, applying it to these interesting tasks is possible.

\subsection{Learning from Hindsight Experience}
Learning from hindsight experience was explored in goal conditioned RL~\citep{kaelbling1993learning, andrychowicz2017hindsight, schaul2015universal}. \citet{andrychowicz2017hindsight} proposes hindsight experience replay (HER) to relabel rewards and transitions retroactively to learn from sparse reward. 
In relation to HER~\citep{andrychowicz2017hindsight}, our work is in the batch setting rather than online setting. 
We propose algorithm improvement to construct hindsight experience directly from offline experience.
HER is designed for Q-learning algorithms~\citep{van2016deep, mnih2013playing, mnih2015human} while \oursabb use next token prediction to learn from hindsight information.
Chain-of-hindsight~\citep{liu2023chain} explores turning all (binary or multi-scale) feedback into a sentence that consists of chain of all feedback and show improve improvements in aligning language models with human preferences.
In relation to it, our work can be seen as applying chain-of-hindsight in the context of automatic feedback. Our work steers model's behavior using the desired target return and reward function at each step as feedback instead of using human preference.

\subsection{Supervised and Meta RL}
Motivated by transforming conventional RL (\eg, policy gradient~\citep{schulman2015trust, schulman2017proximal} and Q-learning~\citep{watkins1989qlearning, mnih2013playing}) as a supervised learning problem, prior work explored various ways~\citep{srivastava2019udrl, paster2020glamor, liu2022masked, carroll2022unimask, chen2021dt, laskin2022context}.
Our work is closely related in that our model is similarly a return conditioned supervised learning. 
At test time, our model can self-improve based upon past experience to try to achieve target desired return.
Using experience to improve model without changing weights is similar to few-shot or in-context learning in large language models~\citep{brown2020language}. 
Recent work Algorithm Distillation (AD)~\citep{laskin2022context} demonstrates similar in-context behaviors in transformer model. 
AD is trained on the lifetime trajectories of a RL algorithm that can solves the task, posing a strong requirement of offline data, while in many important real world domains there exists only diverse, lower return data from multiple sources. 
In relation to AD, \ours can be learned from sub-optimal data by turning the data into \oursdata. 
Leveraging online experience to improve model at test time is related to meta reinforcement learning (meta RL)~\citep{duan2016rl, wang2016learning}. In meta RL the objective is to explicitly optimize for meta learning at test time, while \ours does not, in contrast, the meta learning behavior emerges from training on \oursdata.

\section{Conclusion}
We propose \ours (\oursabb), a Transformer model with the ability of learning by directly combining information from multiple sub-optimal trials and being able to improve itself through multiple trials at test time.
Motivated by prior works on hindsight experience replay and chain of hindsight, the key innovation behind \ours is relabelling multiple trajectories to \oursdata that can be easily constructed from arbitrary offline data.  
On standard RL benchmarks, we showed \oursabb outperforms both strong algorithms designed explicitly for offline RL as well as state-of-the-art Transformer-based policies. 

\textbf{Limitations and Future Work.}
\begin{itemize}[leftmargin=15pt, itemsep=2pt, topsep=2pt]
    \item \emph{Large diverse datasets.} While \ours (\oursabb) outperforms prior transformer-based policies and performs competitively with TD-learning in standard RL benchmarks. \oursabb is a GPT model therefore all limitations of transformer model still apply to \oursabb. For instance, training \oursabb requires large memory because of self-attention quadratic complexity and long sequence length. At test time, rollouting our model is sequential thus slower than non-transformer models. 
    That being said, we believe the advantages of \oursabb outweigh its drawbacks. As we observed in NLP and CV, it is worth scaling transformer-based policies in both model size and dataset size.
    As the datasets used in this work are still small, future work could explore scaling up dataset and model and have more investigation into using large transformer models for RL. 
    \item \emph{Real world applications.} As we observed in the experiments, \ours can learn by directly combining information from multiple sub-optimal trials. Because diverse sub-optimal data is ubiquitous in the real world and \oursabb scales well with model size and dataset diversity, we believe an interesting future direction is applying \oursabb for real-world applications. 
\end{itemize}

\section*{Acknowledgements}
We thank the members of the Berkeley Robot Learning Lab and Berkeley AI Lab for helpful discussions, as well as Google TPU Research Cloud for granting us access to TPUs. 
This project is supported in part by ONR under N00014-21-1-2769.

\bibliography{main}
\bibliographystyle{icml2023}

\newpage
\appendix
\onecolumn
\section{Experimental Details}
The default length of chain of hindsight experience is four (\ie\, input sequence consists of four trajectories) unless mentioned otherwise. 
For small and base model size, we distribute batch size 256 across multiple TPU devices and use gradient accumulation when necessary to reach effective batch size 256. 
For large and x-large model sizes, we distribute model weights across devices and similarly accumulate gradient to reach effective batch size 256.
Our experiments are conducted on TPUv3 32 using Jax and Flax.
On 32 TPUv3, each experiment takes around 4 hours on D4RL and around 6 hours on ExoRL.
Models were trained for $10^5$ gradient steps using the AdamW optimizer.

Our hyperparameters on all tasks are shown below in Table~\ref{tab:hyperparameters}.
In our preliminary experiments on ExoRL, we found that \ours can condition on higher return targets, for fair comparison, we choose the return targets are chosen the same as in prior works. Specifically, on D4RL the target return equals to expert performance for each environment, except for 50\% performance in HalfCheetah, and on ExoRL since the datasets are diverse and contain many lower return trajectories, we choose target returns based on TD3 performance.

\begin{table}[ht]
\caption{Hyperparameters of Agentic Transformers.}
\vskip 0.15in
\begin{center}
\begin{small}
\begin{NiceTabular}{ll}
\toprule
\textbf{Hyperparameter} & \textbf{Value}  \\
\midrule
Number of layers & $3$  \\ 
Number of attention heads    & $1$  \\
Embedding dimension    & $128$  \\
Activation function & ReLU \\
Batch size   & $64$ \\
Dropout & $0.1$ \\
Learning rate & $10^{-4}$ \\
Learning rate decay & Linear warmup for $10^5$ steps \\
Grad norm clip & $0.25$ \\
Weight decay & $10^{-4}$ \\
Initial desired target return at test time (D4RL)   & $6000$ HalfCheetah \\
& $3600$ Hopper \\
& $5000$ Walker \\
Initial desired target return at test time (ExoRL)   & $850$ Walker Stand \\
& $400$ Walker Run \\
& $900$ Walker Walk \\
& $350$ Cheetah Run \\ 
& $300$ Jaco Reach \\ 
& $800$ Cartpole Swingup \\
Number of trajectories to form \oursdata during training & 4 \\ 
Number of trajectories at test time & 4 \\ 
\bottomrule
\end{NiceTabular}
\end{small}
\label{tab:hyperparameters}
\end{center}
\end{table}

\end{document}